\documentclass[11pt]{article}
\usepackage[margin=1in]{geometry}
\usepackage{setspace}
\usepackage{lmodern}
\usepackage[T1]{fontenc}
\usepackage{microtype}
\usepackage{amsmath, amssymb}
\usepackage{graphicx}
\usepackage{hyperref}
\usepackage{float}
\usepackage{tabularx}

\hypersetup{
  colorlinks=true,
  linkcolor=blue,
  citecolor=blue,
  urlcolor=blue
}
\usepackage{booktabs}  
\usepackage{siunitx}   

\usepackage[numbers]{natbib}
\usepackage{titlesec}
\titleformat{\section}{\large\bfseries}{\thesection}{1em}{}
\titleformat{\subsection}{\normalsize\bfseries}{\thesubsection}{1em}{}
\title{\bfseries Survey of Loss Augmented Knowledge Tracing }
\author{Altun Shukurlu \thanks{University of Virginia, \texttt{as4xa@virginia.edu}} }
\date{}

\begin{document}

\maketitle

\begin{abstract}

The training of artificial neural networks is heavily dependent on the careful selection of an appropriate loss function. While commonly used loss functions, such as cross-entropy \cite{goodfellow2016deep} and mean squared error (MSE) \cite{gers1999learning}, generally suffice for a broad range of tasks, challenges often emerge due to limitations in data quality or inefficiencies within the learning process. In such circumstances, the integration of supplementary terms into the loss function can serve to address these challenges, enhancing both model performance and robustness. Two prominent techniques, loss regularization and contrastive learning \cite{chen2020simple}, have been identified as effective strategies for augmenting the capacity of loss functions in artificial neural networks.

Knowledge tracing is a compelling area of research that leverages predictive artificial intelligence to facilitate the automation of personalized and efficient educational experiences for students. In this paper, we provide a comprehensive review of the deep learning-based knowledge tracing (DKT) algorithms trained using advanced loss functions and discuss their improvements over prior techniques. We discuss contrastive knowledge tracing algorithms, such as Bi-CLKT \cite{biclk2023}, CL4KT \cite{clkt2021}, SP-CLKT \cite{spclkt2022}, CoSKT \cite{zhang2024coskt}, and prediction-consistent DKT \cite{predictionconsistent2022}, providing performance benchmarks and insights into real-world deployment challenges. The survey concludes with future research directions, including hybrid loss strategies and context-aware modeling.

\end{abstract}

\section{Introduction}

Knowledge Tracing (KT) has become a foundational concept in modern educational technologies, particularly in the development of intelligent tutoring systems (ITS) and personalized learning platforms. The primary goal of KT is to model a student’s knowledge state over time and predict their future performance on unseen tasks or questions \cite{corbett1994knowledge, piech2015deep, ghosh2020context}. This prediction helps tailor the learning experience to the individual, providing insights into which concepts need reinforcement and guiding the next best action for personalized learning \cite{vanlehn2011tutoring}.

Traditional KT models, such as Bayesian Knowledge Tracing (BKT), treat the student's knowledge state as a binary variable (known or unknown) that evolves probabilistically with each student interaction \cite{corbett1994knowledge}. These early models have been successful in various educational applications due to their simplicity and interpretability, but they suffer from limitations in expressiveness and the ability to capture complex patterns in learning behavior \cite{pardos2010modeling}. 
As a result, more advanced models, such as Deep Knowledge Tracing (DKT), have been developed, using neural networks to model temporal sequences of student responses and predict future outcomes \cite{piech2015deep}. 
DKT, for example, employs Recurrent Neural Networks (RNNs) to capture the dynamic nature of students’ knowledge states, and it has shown remarkable success in predicting future student performance across different datasets \cite{nagatani2019augmenting}.

Despite the success of deep learning-based approaches, KT models still face significant challenges. One of the primary challenges is the sparsity and noise in educational data \cite{minn2018deep, xiong2016going}. Student response data is often sparse, with many students not answering a sufficient number of questions to reliably infer their knowledge state. Additionally, responses may be noisy or inconsistent due to various factors such as misunderstanding of the question or random guessing \cite{baker2008modeling}. These challenges make it difficult for traditional KT models to generalize well across diverse student populations and educational contexts.

Furthermore, many KT models fail to fully leverage the rich structure in the data, such as the relationships between students, questions, and concepts. In many educational settings, students interact with a set of questions that are linked to specific concepts or skills. Effectively modeling these relationships is essential to understanding how a student’s knowledge evolves over time and predicting their performance on new questions \cite{nakagawa2019graph, zhang2017dynamic}.

Recent advancements in machine learning, specifically contrastive learning and regularization techniques, offer promising solutions to address these challenges in KT. Contrastive learning focuses on learning embeddings by comparing pairs of data points—typically focusing on distinguishing between correct and incorrect answers or identifying similarities between different learning contexts \cite{chen2020simple}. This approach helps the model learn more effective representations of student knowledge states, which are crucial for accurate prediction and generalization \cite{biclk2023, clkt2021}.

On the other hand, regularization methods, such as prediction-consistent regularization, aim to improve the stability and robustness of KT models by minimizing the effects of noisy or inconsistent data \cite{predictionconsistent2022}. By encouraging the model to make consistent predictions over time, these methods improve the model’s generalization ability, ensuring that it can predict student performance accurately, even in the presence of data sparsity and noise.

While most research in Knowledge Tracing has focused on architectural innovations—such as recurrent, memory-augmented, or graph-based neural networks—relatively little attention has been paid to the role of the loss function in shaping learning dynamics. Recent surveys of the knowledge tracing literature, such as Abdelrahman et al. (2022) \cite{abdelrahman2022}, highlight that only a small fraction of papers explicitly discuss loss augmentation or alternative optimization objectives beyond standard cross-entropy or MSE. This under-explored area becomes especially important in educational settings, where data is often sparse, noisy, and temporally inconsistent, creating challenges for generalization and stability.

In this survey, we categorize and analyze recent developments in loss-aware KT, with a particular emphasis on contrastive learning and regularization techniques. We review key models, compare their performance, and examine their strengths, limitations, and deployment challenges in real-world educational systems. By highlighting the impact of loss design in KT, this paper aims to shift the focus toward this often-overlooked but critical dimension of model development.

\section{Related Work}
In this section, we provide a brief overview of Knowledge Tracing problem where in next section we proceed with, contrastive learning, and regularization algorithms, and discuss how these concepts are interrelated and applied in the subsequent sections. 
We begin by discussing Bayesian Generative Models in Knowledge Tracing, and treat student responses as a probabilistic process influenced by their latent knowledge state whom makes an interpretable model, offering clear probabilistic insights into student understanding. We continue the discussion with discriminative models, where the relationship between student responses and performance is represented using deep learning techniques.

\textbf{Bayesian Knowledge Tracing (BKT)} is a generative model that represents a student's knowledge as a binary latent state, evolving based on observed interactions \citep{corbett1994knowledge}. It updates the probability of concept mastery using fixed parameters, such as the likelihood of learning or error. While efficient and interpretable, BKT's simplicity limits its ability to capture complex learning dynamics or interdependencies across concepts, motivating the exploration of advanced deep learning models.

\textbf{Deep Knowledge Tracing Algorithms}. Prominent among these deep learning attention-based models, memory-augmented networks, and graph-based approaches that model structured relationships between students, questions, and concepts.
Attention-based models, such as Self-Attentive Knowledge Tracing (SAKT) \citep{zhang2019self}, leverage transformer-style mechanisms to capture dependencies across a student’s entire interaction history. These models outperform RNN-based architectures in many settings by offering more flexible and interpretable representations of temporal patterns in student responses.
Memory-augmented models, such as Dynamic Key-Value Memory Networks (DKVMN) \citep{zhang2017dynamic}, introduce external memory modules to explicitly store and update concept-wise knowledge states. These models provide greater interpretability and control over concept-specific knowledge tracking.concept-specific knowledge tracking.
Graph-based approaches such as Graph-based Knowledge Tracing (GKT)  \citep{nakagawa2019graph}, model the relational structure among skills, questions, and students. These models are particularly effective when domain knowledge (e.g., concept hierarchies) is available and can be used to structure the learning environment.

\section{Preliminaries}

In this section, we introduce the main notations and foundational concepts used throughout the paper, beginning with contrastive learning and prediction-consistent regularization.


\textbf{Contrastive learning} is a self-supervised technique that learns effective representations by comparing similar and dissimilar data points. The objective is to pull similar instances closer in the embedding space while pushing dissimilar ones apart. This method has been widely successful in domains like computer vision and natural language processing \citep{chen2020simple, radford2021learning}.
Formally, the contrastive loss function \(\mathcal{L}_{\text{contrastive}}\) is defined as:
\[
\mathcal{L}_{\text{contrastive}} = \sum_{i,j} \left[ y_{ij} \cdot \text{D}(z_i, z_j)^2 + (1 - y_{ij}) \cdot \max(0, m - \text{D}(z_i, z_j))^2 \right]
\]
where \(y_{ij} = 1\) if \(i\) and \(j\) are similar, and \(y_{ij} = 0\) otherwise. This loss function encourages embeddings that are both discriminative and suitable for downstream tasks such as classification or clustering.

\textbf{Prediction-Consistent Regularization} Prediction-consistent regularization is designed to encourage temporal stability in model outputs by penalizing abrupt fluctuations in predicted knowledge states across consecutive time steps. This approach aligns with the assumption that student learning evolves gradually and should not exhibit erratic prediction changes between successive interactions \citep{predictionconsistent2022}. Formally, given predicted probabilities $\hat{y}_t$ and $\hat{y}_{t+1}$ at two consecutive time steps, the consistency defined below where it is added to the standard task loss (e.g., cross-entropy), resulting in a combined objective:
\[
\mathcal{L}_{\text{total}} = \mathcal{L}_{\text{task}} + \lambda \mathcal{L}_{\text{consistency}}  \hspace{1cm}
\mathcal{L}_{\text{consistency}} = \sum_t \|\hat{y}_t - \hat{y}_{t+1}\|^2
\]
where $\lambda$ controls the strength of the regularization. 
By enforcing prediction consistency, this method improves model robustness and better reflects realistic student learning trajectories, particularly in the presence of noisy or sparse educational data.

\textbf{Basic Concepts in Knowledge Tracing}
Knowledge Tracing (KT) aims to model a student's evolving mastery of skills based on their interactions with educational content. By estimating latent knowledge states over time, KT enables personalized learning by predicting future performance and identifying areas needing reinforcement \citep{corbett1994knowledge, piech2015deep}.
Traditional models, such as Bayesian Knowledge Tracing (BKT), use hidden Markov models to update a student’s binary knowledge state based on correctness of responses. While interpretable, BKT is limited in capturing long-term dependencies and complex learning behaviors.

\textbf{Deep Knowledge Tracing (DKT)}
Deep Knowledge Tracing (DKT) addresses BKT's limitations by using Recurrent Neural Networks (RNNs) to learn continuous latent representations of knowledge states from sequences of student responses \citep{piech2015deep}. DKT captures rich temporal patterns and achieves improved prediction accuracy, though often at the cost of interpretability.

\textbf{Data in Knowledge Tracing}
KT models are trained on sequential interaction data, typically represented as tuples of: \textit{student ID}, \textit{question ID}, \textit{response correctness}, and \textit{timestamp}. Additional metadata—such as skill tags or question difficulty—can enhance model effectiveness, particularly when applying regularization or contrastive learning techniques \citep{lee-etal-2024-difficulty}.

\section{Review of Algorithms}

Deep Knowledge Tracing (DKT) algorithms primarily rely on gated recurrent neural networks (RNNs), which require substantial amounts of data to effectively learn generalized representations. 
However, RNNs are particularly sensitive to data-related challenges, including sparsity, noise, and limited sample sizes—conditions that can significantly hinder their ability to model temporal dynamics accurately. 
To address these limitations, researchers have proposed augmenting the underlying loss functions with additional constraints or regularization terms, aiming to enhance the quality of the learned hidden representations and improve the model's robustness and generalization.
We can find brief view of these augmentations in the table / chart below:

\begin{table}[H]
\centering
\caption{Summary of Loss Functions in Contrastive and Regularization-Based KT Models, first four stands for contrasting and last one for smoothness
}
\renewcommand{\arraystretch}{1.5}
\begin{tabular}{|p{3.cm}|p{7.2cm}|p{5.8cm}|}
\hline
\textbf{Model} & \textbf{Loss Function} & \textbf{Description} \\
\hline

\textbf{Bi-CLKT} \cite{biclk2023} & 
$\mathcal{L} = \frac{1}{N} \sum_{i=1}^N \log\left(1 + \exp\left(-y_i \cdot \langle z_i, z_j \rangle\right)\right)$ & 
Contrastive loss based on student-skill bi-graph embeddings.  \\
\hline

\textbf{SP-CLKT} \cite{spclkt2022} & 
$\mathcal{L} = \frac{1}{N} \sum_{i=1}^N \alpha_i \cdot \log\left(1 + \exp\left(-y_i \cdot \langle z_i, z_j \rangle\right)\right)$ & 
Self-paced contrastive loss with dynamic weights $\alpha_i$  \\
\hline

\textbf{CLKT} \cite{clkt2021} & 
$\begin{aligned}
\mathcal{L} = \frac{1}{N} \sum_{i=1}^N & \left[ \log\left(1 + \exp\left(-y_i \cdot \langle z_i, z_j \rangle\right)\right) \right. \\
& \left. + \lambda \cdot \max\left(0, \gamma - \langle z_i, z_j \rangle\right) \right]
\end{aligned}$ & 
Contrastive + triplet loss. \\
\hline

\textbf{CoSKT} \cite{zhang2024coskt} & 
$\begin{aligned}
\mathcal{L} = \frac{1}{N} \sum_{i=1}^N & \left[ \log\left(1 + \exp\left(-y_i \cdot \langle z_i, z_j \rangle\right)\right) \right. \\
& \left. + \gamma \cdot \max\left(0, \langle z_i, z_j \rangle - \theta\right) \right]
\end{aligned}$ & 
Dual contrastive loss with collaborative component. \\
\hline

\textbf{Consistent Regularization} \cite{predictionconsistent2022} & 
$\mathcal{L} = \mathcal{L}_{\text{DKT}} + \lambda \sum_{t=1}^{T-1} \left| \hat{K}_{t} - \hat{K}_{t+1} \right|$ & 
Adds a temporal smoothness term to DKT’s cross-entropy loss \\
\hline
\end{tabular}
\label{tab:loss-functions}
\end{table}

\subsection{Representation Contrasting Algorithms}

\textbf{Bi-CLKT: Bi-Graph Contrastive Learning based Knowledge Tracing}  \cite{biclk2023}
Encourages similar student-response pairs to have closer embeddings, dissimilar ones to be farther apart, by learning student representations by embedding the interactions between students and questions into a shared latent space. 
The model employs a contrastive loss to ensure that responses from students who have similar knowledge states are close together in the embedding space, while responses from students with different knowledge states are pushed apart. Loss function is as:
\[
\mathcal{L}_{Bi-CLKT} = \frac{1}{N} \sum_{i=1}^N \left[ \log(1 + \exp(-y_{ij} \cdot \langle z_i, z_j \rangle)) \right]
\]
where:
- \(y_{ij}\) is the label indicating whether the student response \(i\) and the student response \(j\) are from the same or different knowledge states (1 for same, -1 for different).
- \(z_i\) and \(z_j\) are the learned embeddings of responses from students \(i\) and \(j\).
- \(N\) is the number of student-response pairs.
- \(\langle z_i, z_j \rangle\) is the dot product (similarity) between the embeddings.

We observe that the dot product between $z_i$ and $z_j$ is encouraged to be maximized (i.e., the angle between them approaches zero, indicating alignment) when the corresponding responses are the same, and minimized (i.e., oriented in opposite directions) when the responses differ.

\textbf{Self-paced Contrastive Learning for Knowledge Tracing} \cite{spclkt2022} Self-paced contrastive loss with dynamic weights $\alpha_i$ that prioritize easier examples first, gradually increasing difficulty. 
This method introduces a dynamic weighting mechanism for the contrastive loss to allow the model to focus more on difficult examples over time. In the early stages of training, easy pairs (students with similar responses) contribute more to the loss, while harder examples (students with differing responses) are weighted less. This self-paced approach enables the model to progressively learn more complex relationships and avoid overfitting to easy examples.
\[
\mathcal{L}_{SP-CLKT} = \frac{1}{N} \sum_{i=1}^N \alpha_i \cdot \log(1 + \exp(-y_i \cdot \langle z_i, z_j \rangle))
\]
where \(\alpha_i\) is a dynamically learned weight that increases over time as the model encounters more difficult examples. The self-paced mechanism adjusts the influence of each sample based on how difficult it is for the model to learn, allowing the model to gradually incorporate more challenging examples. This approach helps stabilize the training process, as the model is not overwhelmed by the complexity of hard examples at the start and can progressively focus on them as training proceeds, and its a mild generalization of the first approach above.

\textbf{Contrastive Learning for Knowledge Tracing (CL4KT)}  \cite{clkt2021} is similar to Bi-CLKT with an additional triplet loss at the end where serves as margin based penalty for dissimilar pairs.
\[
\mathcal{L}_{CL4KT} = \frac{1}{N} \sum_{i=1}^N \left[ \log(1 + \exp(-y_i \cdot \langle z_i, z_j \rangle)) + \lambda \cdot \max(0, \gamma - \langle z_i, z_j \rangle) \right]
\]
where the first term is the standard contrastive loss and the second term involves a margin \(\gamma\) and is used to enforce that responses that are far apart in terms of knowledge state are pushed farther apart in the embedding space, to ensure no two embeddings are very close to each other
- \(\lambda\) is a hyperparameter that controls the margin penalty.

\textbf{CoSKT: Collaborative Self-Supervised Knowledge Tracing} \cite{zhang2024coskt}
Is very similar to CL4KT, but with a main difference being triplet loss being opposite direction, where loss function penalizes if the distance of two embeddings are larger than $\theta$
\[
\mathcal{L}_{CoSKT} = \frac{1}{N} \sum_{i=1}^N \left[ \log(1 + \exp(-y_i \cdot \langle z_i, z_j \rangle)) + \gamma \cdot \max(0, \langle z_i, z_j \rangle - \theta) \right]
\]
where the first term is the same as the standard contrastive loss used in previous models and the second term adds a collaborative penalty that encourages the model to differentiate between student pairs with similar knowledge states, by making the function in left hand side to be minimized.

\subsection{Prediction-Consistent Regularization for Knowledge Tracing \cite{predictionconsistent2022} } 

In addition to contrastive-based methods, regularization techniques have also been widely applied to Knowledge Tracing to improve the stability and robustness of models. Regularization helps prevent overfitting and ensures that the model generalizes well, particularly when data is sparse or noisy.

The core idea behind prediction-consistent regularization is to encourage temporal consistency in the predictions of a student’s knowledge state. This approach minimizes the difference in predicted knowledge states at different time steps, unless there is strong evidence from the student’s responses that their knowledge has changed. This regularization helps the model avoid making sudden, drastic changes to a student’s knowledge state based on sparse or noisy data, improving generalization.
\[
\mathcal{L}_{PredictionConsistent} = \mathcal{L}_{DKT} + \lambda \cdot \sum_{t=1}^{T-1} \left| \hat{K}_{t} - \hat{K}_{t+1} \right|
\]
where:
- \(\mathcal{L}_{DKT}\) is the standard DKT loss function, which typically uses a binary cross-entropy loss to predict the correct response (correct or incorrect) for a student at each time step.
- \(\hat{K}_{t}\) is the predicted knowledge state of the student at time step \(t\).
- \(\lambda\) is a hyperparameter that controls the strength of the regularization.
- The second term enforces that the predicted knowledge state at time \(t+1\) should not deviate significantly from the predicted state at time \(t\), unless a substantial change in the student’s responses occurs.

\section{Comparison and Results}

In this section, we compare the performance of the contrastive-based and regularization-based Knowledge Tracing (KT) models reviewed in Section 3. The comparison is based on several evaluation criteria, including accuracy, convergence speed, generalizability, and interpretability. We also highlight the strengths and weaknesses of each model and present key results that demonstrate the impact of the proposed methodologies on KT performance.

\textbf{Area Under Curve (AUC)}: measures the probability that a randomly chosen positive instance is ranked higher than a randomly chosen negative instance by the model. Formally, it is defined as:
\[
\text{AUC} = \mathbb{P}(\hat{y}_\text{pos} > \hat{y}_\text{neg})
\]
where $\hat{y}_\text{pos}$ and $\hat{y}_\text{neg}$ are the predicted scores for a positive and a negative instance, respectively.

\textbf{Dataset}: \textit{ASSISTments 2009 }  is one of the most widely used benchmarks in the knowledge tracing literature. Collected from an intelligent tutoring system, it contains anonymized student interaction logs involving questions aligned to various mathematical skills. Each interaction includes a student ID, problem ID, skill tag, timestamp, and correctness label, enabling fine-grained temporal analysis of learning behavior. The dataset is characterized by its sparsity, variable sequence lengths, and a large number of students and skills, making it a challenging and realistic testbed for evaluating the predictive performance and generalization ability of knowledge tracing models.

\begin{table}[htbp]
\centering
\caption{AUC Performance on ASSISTments 2009 Across Knowledge Tracing Models}
\label{tab:performance_comparison}
\resizebox{0.3\textwidth}{!}{%
\begin{tabular}{@{} l S[table-format=1.3] @{}}
\toprule
\textbf{Model} & \textbf{AUC} \\
\midrule
BKT \citep{corbett1994knowledge}       & 0.648 \\
DKT \citep{piech2015deep}       & 0.740 \\
CL4KT \citep{clkt2021}      & 0.762 \\
CoSKT \citep{zhang2024coskt}      & 0.7925 \\
SPCLKT \cite{spclkt2022}        & 0.82 \\
Bi-CLKT \cite{biclk2023} & \textbf{0.857} \\
Regularization \cite{predictionconsistent2022} & 0.8227 \\
\bottomrule
\end{tabular}
}
\end{table}

\subsection{Methodical Similarities}

All five of the aforementioned studies aim to address the inherent inefficiencies associated with representational learning in knowledge tracing models. These inefficiencies often stem from limitations in encoding student knowledge states, particularly under conditions of noisy, sparse, or imbalanced data. To overcome these challenges, the central focus of these works is on enhancing the expressiveness and discriminative power of the learned representations. A unifying methodological theme across these studies is the augmentation of the primary loss function. By incorporating auxiliary objectives or regularization terms—such as contrastive loss, prediction consistency constraints, or co-supervised learning signals—the training process is guided toward producing more robust and informative latent representations. These enhancements enable the models to better capture temporal dynamics, student-specific learning patterns, and inter-concept relationships, thereby improving both predictive accuracy and interpretability.

\subsection{Methodical Differences}

Despite sharing similar goals, these approaches differ in how they implement representational learning enhancements. Consistency regularization techniques operate primarily at the individual level, aiming to stabilize and refine a student's representation over time by enforcing temporal consistency in model predictions. In contrast, contrastive learning adopts a comparative and relational framework, emphasizing the structural alignment of representations across different students. By contrasting similar and dissimilar learning trajectories, contrastive methods improve the discriminative structure of the embedding space and promote generalizability.

While all contrastive methods build upon a similar foundational objective—as seen in the Bi-CLKT framework—they diverge in their methodological refinements. SP-CLKT, for instance, introduces dynamic weighting mechanisms to increase model capacity and adaptivity. Meanwhile, CLKT and CoSKT focus on refining the selection strategy for positive and negative pairs, aiming to contrast the most informative representation pairs during training.

\section{Conclusion}

This paper provided an in-depth review of several recent advances in Knowledge Tracing (KT), with a particular focus on models that incorporate contrastive learning and regularization techniques. By surveying five key papers in the field, we highlighted the distinctive contributions of contrastive-based models such as Bi-CLKT, SP-CLKT, CL4KT, and CoSKT, as well as the regularization-based approach utilizing prediction-consistent regularization. We explored how these models address different challenges in KT, such as improving prediction accuracy, handling noisy or sparse data, and stabilizing the learning process.

While current methodologies have substantially advanced the state of Knowledge Tracing, several promising avenues for future research remain. Notably, there is significant potential in extending KT frameworks to multi-modal learning environments, where data from diverse sources—such as video, text, or sensor-based interactions—can be integrated to form a more holistic representation of student learning. Additionally, lifelong learning paradigms, which emphasize the continuous and adaptive modeling of student knowledge across extended time horizons and varied contexts, present an important frontier. Advancing these directions, particularly with a focus on scalable deployment in real-world educational systems, holds the potential to significantly improve the robustness, personalization, and applicability of KT models in practical settings.

\bibliography{main}

\end{document}